# Advantages of Machine Learning in Bus Transport Analysis

Amirsadegh Roshanzamir
*M.Sc Graduated,*
*Department of Electrical Engineering,*
*Sharif University of Technology*
*roshanzamir_as@alum.shaif.edu,*

**ABSTRACT**
Supervised Machine Learning is an innovative method that aims to mimic human learning by using past experiences. In this study, we utilize supervised machine learning algorithms to analyze the factors that contribute to the punctuality of Tehran BRT bus system. We gather publicly available datasets of 2020 to 2022 from Municipality of Tehran to train and test our models. By employing various algorithms and leveraging Python's Sci Kit Learn and Stats Models libraries, we construct accurate models capable of predicting whether a bus route will meet the prescribed standards for on-time performance on any given day. Furthermore, we delve deeper into the decision-making process of each algorithm to determine the most influential factor it considers. This investigation allows us to uncover the key feature that significantly impacts the effectiveness of bus routes, providing valuable insights for improving their performance.

---

## I. INTRODUCTION

The rise of the internet and technology has revolutionized the way we record and share information. We now have an overwhelming amount of data at our fingertips, making it nearly impossible for humans to extract meaningful insights efficiently. However, machines have emerged as valuable allies in this data-driven era, helping us make sense of the vast information available. While humans can often make accurate judgments, quantifying the decision-making process can be challenging. This is where supervised machine learning steps in.

Supervised machine learning algorithms aim to provide a logical explanation for why something is the way it is. This approach closely resembles how humans learn. For example, when a child is shown pictures of kittens and told by their parent that those images represent kittens, the child gradually learns to correctly identify a kitten when they see one. Humans can classify new instances based on past experiences. Supervised machine learning seeks to replicate this behavior by manually labeling a subset of a dataset and using computers to recreate this experience-based learning process. Our goal is to apply established supervised machine learning techniques to solve a unique problem.

## II. MACHINE LEARNING IN TRANSPORTATION

Machine learning algorithms have been widely used in the field of public transportation to address issues such as bus punctuality. These algorithms have been employed to predict variables like bus arrival times, travel durations, and dwell times. With advancements in data collection and storage technologies, machine learning has become an essential tool in analyzing the performance of public transit systems.

For example, in Seoul, South Korea, a non-parametric regression method called nearest neighbor has been used to analyze bus travel time [2]. This algorithm considers historical and current path travel times of neighboring paths to predict bus travel time. Compared to the traditional plain historical average technique, the nearest neighbor algorithm has shown superior accuracy, especially during daytime when travel time fluctuations are more pronounced.

Another aspect studied in public transportation is bus bunching, where buses tend to bunch together instead of maintaining a regular spacing. Factors such as the day of the week, bus dwell time, intersection delay, schedule deviation, bus spacing, and proximity to bus stops are used to model this problem [1]. Machine learning algorithms like Gene Expression Programming and Decision Tree have been employed to identify the most influential factors in bus bunching. The Decision Tree algorithm has emerged as the most accurate, with the day of the week identified as the most significant factor.

Overall, machine learning algorithms have proven to be effective in analyzing and improving various aspects of public transportation systems worldwide. These algorithms leverage the vast amount of transportation data available and provide reliable and efficient solutions for enhancing the performance of public transit systems and here in this paper we want to deal with bus on-time problem in Tehran.

## III. PROBLEM EXPLANATION

This study examines the timeliness of Bus Rapid Transport (BRT) bus routes in Tehran which it's mapped depicted in Fig. 1. As public transportation is relied upon by many individuals, it is crucial for buses to adhere to their schedules. On-time performance can be measured by calculating the proportion of times a bus arrives on time out of the total number of stops. However, there are additional factors within the bus system that, while not directly affecting this calculation, still impact punctuality. For example, the geographical location within the city or the duration a bus waits for passengers may be correlated with a bus's punctuality.

### A. Dataset

To obtain a suitable dataset for this study, we utilized the data from 2020 to 2022 of Tehran which are provided by Municipality of Tehran. This organization collected data on various city services, including electricity prices and the SA aquifer authority. The chosen dataset is the Tehran bus on-time performance dataset, which contains data recorded by the buses themselves while operating on specific routes. We modified the dataset based on its relevant map to incorporate location information. The columns provide details about each route, such as the frequency of buses being early, on time, late, and more. Each row represents the cumulative data of all buses on a particular route throughout one day. The dataset consists of more than 10,000 rows, representing the daily data of each route from March 20th, 2020 (which is the first day of Iranian calendar) to March 20th, 2022, covering a period of two years. This study aims to utilize this dataset to determine the significance of all included factors and more in relation to their impact on the punctuality of bus routes.

One of the most challenging aspects of any machine learning project is collecting, cleaning, and enhancing a dataset. In this case, collection was straightforward as we utilized publicly available data. The dataset used for this study was relatively clean. If a row lacked certain data due to errors or issues with the tracking equipment, the on-time performance column would be marked as 'NULL'. This facilitated the cleaning process, and we decided to clean the dataset during runtime by removing rows with a 'NULL' value in the on-time performance column. In terms of enhancements, our final dataset for this project differs slightly from the initial download from the Municipality of Tehran website. The first modification we made was adding an additional column called "bus_target". This column is binary, with a value of 1 indicating a bus route with an on-time performance of 95% or higher, and a value of 0 indicating a lower performance. This was primarily done because most regression algorithms exhibited poor accuracy.

By converting the problem into a classification task, we were able to utilize classification algorithms for the same objective. This allowed us to assess the regression algorithms that displayed high accuracy in predicting on-time performance, while also employing classification algorithms for the new target variable. Also, to enhance the dataset, we included geographical location by introducing binary columns for north, south, east, and west. This was done manually by referring to the city map of bus routes (Fig. 1.) and dividing the city into four sections using an X. A bus route falling within a particular section was classified accordingly. If a route crossed over into another section significantly, it was considered part of that section as well. For example, if a route traversed half of the north section and half of the east section, it would be assigned a value of 1 in both the north and east columns.

**B. Machine Learning Algorithms**

Initially, we focused on the decision tree algorithm and its variant, random forest. These served as our baseline for comparing the performance of other algorithms. Random forest was chosen to validate our decision tree results. Subsequently, we explored the application of Naive Bayes to our dataset. Given the size of our data and the number of categories involved, we believed this algorithm would be suitable. We then experimented with K-Nearest Neighbor, as it aligned with our requirements as a member of the nearest neighbor family. Logistic regression was another valuable algorithm we employed to cross-check outcomes from other algorithms. Finally, we delved into Poisson regression, which proved to be a more complex algorithm. Our goal was to determine its feasibility and whether it could provide meaningful insights from our data.

## IV. ANALYSIS AND RESULTS

**A. Decision Tree**

Decision trees are widely used in classification problems, and they have proven to be quite useful in this study as well. One of the advantages of using decision trees is their ability to incorporate regression methods and handle multiple variables simultaneously. However, it is important to consider the disadvantages of this algorithm. One major drawback is that decision trees can quickly become complex, making it difficult to interpret the results and analyze the data. In our study, we encountered this issue, which hindered our analysis. Additionally, small changes in starting conditions can lead to significant changes in the results, further complicating the analysis [3].

Despite these challenges, the decision tree classifier demonstrated high accuracy right from the start. On average, decision trees achieved 88.5% accuracy in various runs. However, the initial tree produced was large and difficult to comprehend due to its sheer size. To address this issue, we reduced the maximum size of the tree to four levels, as shown in Fig. 2. Although this reduction in size resulted in a decrease in accuracy to 83%, it significantly improved readability. In Fig. 2, orange boxes represent instances of poor performance, while green boxes represent instances of good performance. The darkness of the color indicates the level of confidence in a prediction.

In a decision tree, the most important features are displayed higher up in the tree. In our dataset, the most crucial factor identified by the decision tree was late arrival. Since the algorithm did not have sufficient data to directly calculate on-time performance, it inferred that buses were falling behind relatively early in the route when late arrivals occurred. This delay then compounded at subsequent stops. To gain further insights, we generated a new decision tree by removing late arrivals from the dataset, as depicted in Fig. 3.

Fig. 3 illustrates that routes below or above 1044 are the most significant factors at the top of the tree. This finding is important because routes above 1044 are the newest routes. On the right side of the tree (which in our figure is depicted in lower part of figure), which represents the newer routes, we observe better performance. Although the tree in Figure 3 is limited for human readability and does not have a box indicating good performance on the right side, it is clear that the prediction of "Bad Performance" is less confident compared to the left side. This suggests two possibilities: 1) the new routes are more efficient, and 2) they experience less traffic or a combination of both. It would be worthwhile to investigate the factors contributing to the performance of these new routes and potentially consider redesigning older routes based on those factors.

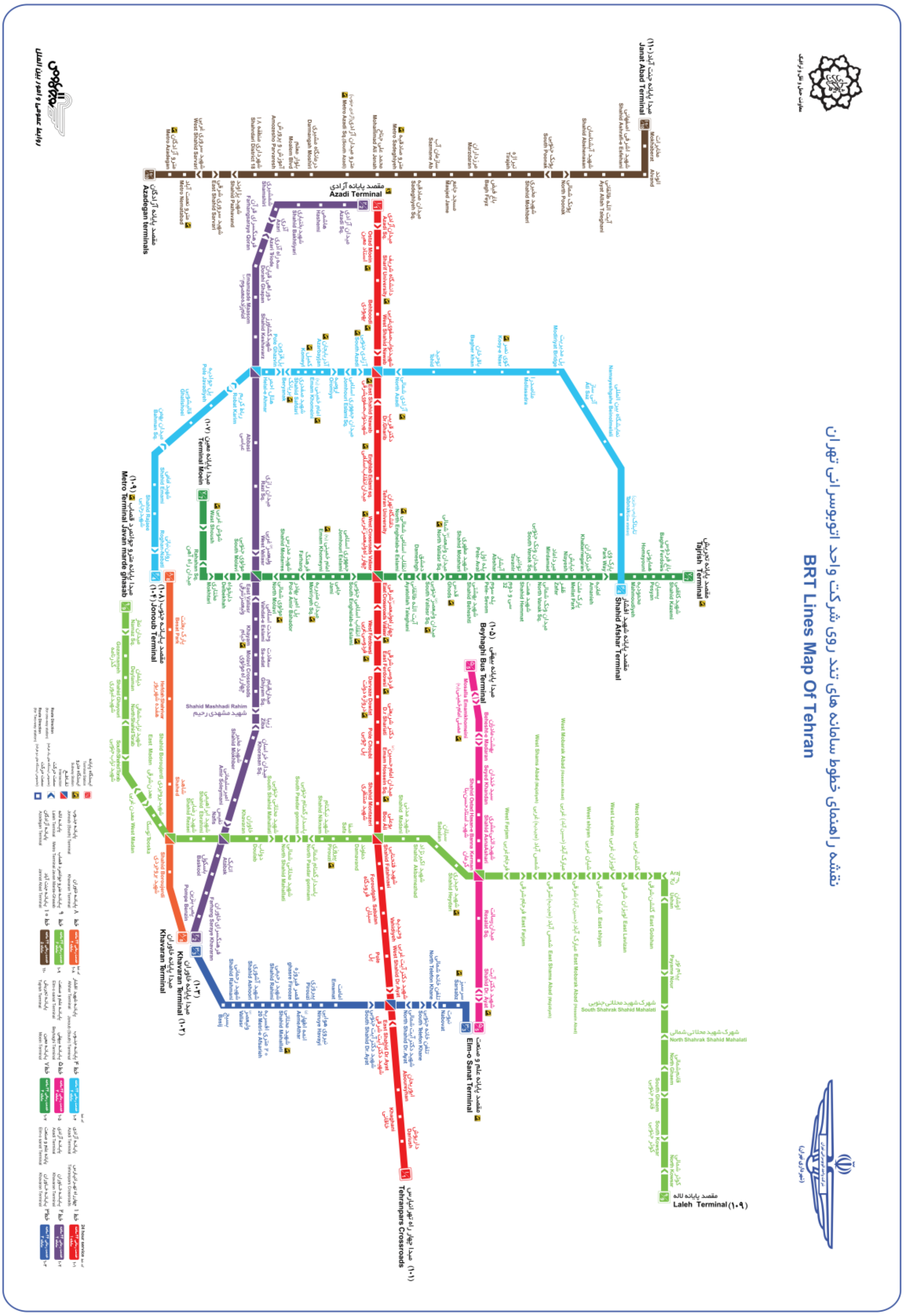

**Fig. 1. Tehran, BRT Lines**

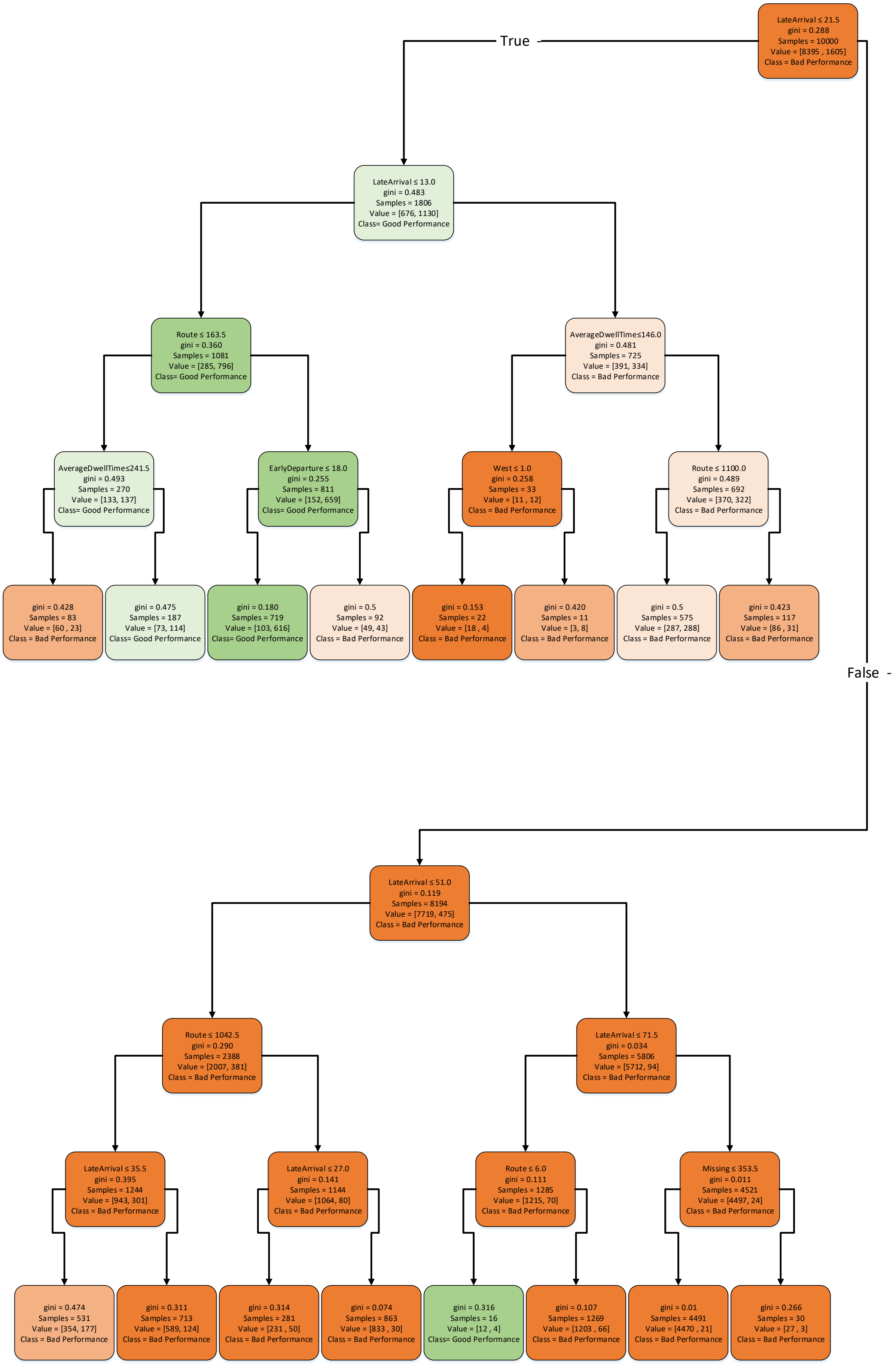

**Fig. 2. Machine Learning Decision Tree**

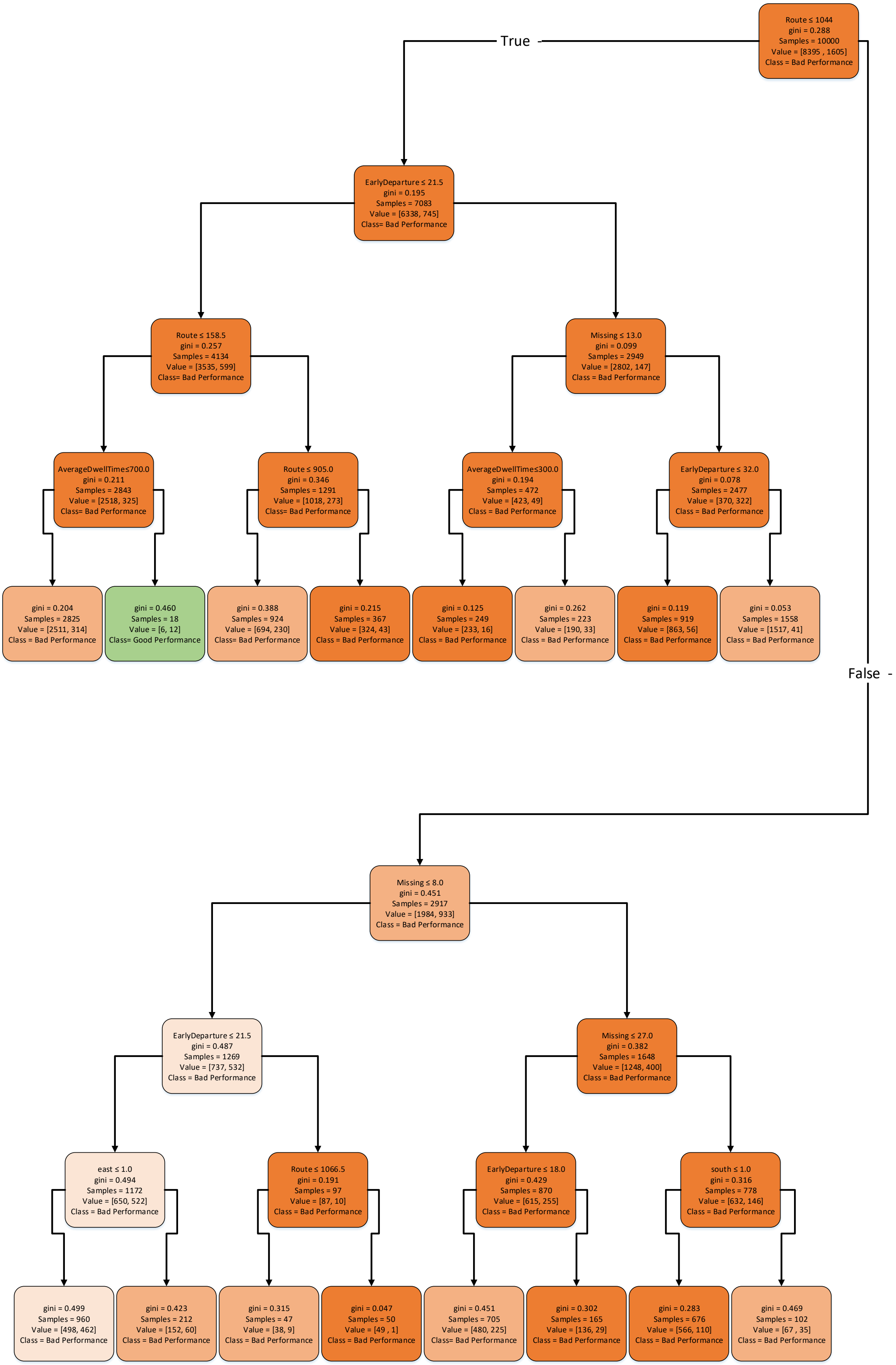

**Fig. 3. Machine Learning Decision Tree without Late Arrival**

**B. Naive Bayes**

The Naive approach assumes that the variables involved are independent, allowing it to achieve good results with smaller data sets compared to other machine learning algorithms. However, it is also scalable and useful for large data sets. However, if the variables are not truly independent, it can lead to erroneous results. Additionally, data often needs to be smoothed and less frequently used categories may need to be discarded in order to obtain accurate predictions [5].

In contrast, the Bayesian algorithm did not provide useful results for the problem at hand. The results were heavily skewed towards the negative end and further manipulation did not significantly improve them. This contradicted other findings and suggests that the algorithm is not valid for this particular data set. It is reasonable to conclude that the variables in this case were too dependent on each other for the algorithm to work effectively. The assumption of naive conditional independence equation (1) states that:

$$P(x_i|y, x_1, \dots, x_{i-1}, x_{i+1}, \dots, x_n) = P(x_i|y) \quad (1)$$

Therefore, a straightforward calculation, as shown in Table 1, reveals that this underlying assumption is violated by the data, which aligns with human intuition. The pandas package offers a convenient way to calculate the statistical pair-wise correlation using dataframe.corr().

**TABLE 1. STATISTICAL CORRELATION BETWEEN PAIR-WISE VARIABLES**

| | Early Departure | Late Arrival | Missing | Dwell Time |
|---|---|---|---|---|
| Early Departure | 1.00 | 0.27 | 0.19 | -0.07 |
| Late Arrival | 0.27 | 1.00 | 0.41 | -0.13 |
| Missing | 0.19 | 0.41 | 1.00 | -0.02 |
| Dwell Time | -0.07 | -0.13 | -0.02 | 1.00 |

There is a significant correlation between Late Arrival and Early Departure, which breaks the underlying assumption of Naive Bayes algorithms. While it is true that Naive Bayes can still be successful in dealing with many real-world problems even when variables are not independent, it fails to outperform other algorithms in the case of Tehran bus on-time performance. The main reason for this can be attributed to the strong correlation between variables.

**C. Random Forest**

The Random Forest technique, while it may not offer direct insights on its own, holds great significance as a model. However, one drawback of the Sci Kit Learn random forest models is their inability to combine classifiers with regressors. As a result, we can only compare the group of classifiers, which happens to be the largest group of algorithms we use. The underlying idea is that if the random forest can enhance accuracy, the algorithm with the highest weight becomes more important than the others. Nevertheless, this does not mean that the other algorithms should be ignored. Randomization plays a crucial role in this approach, making it suitable for large datasets where randomization evens out [4].

The main goal of the random forest is to compensate for weaknesses and improve accuracy. By conducting various experiments with different weights to find the optimal outcome, we were able to achieve an average accuracy of 90.73%. This impressive accuracy was achieved by assigning the highest weights to the decision tree and K nearest neighbor algorithms. The fact that this random forest model outperforms each individual algorithm further strengthens the idea that these algorithms are the most dependable for decision-making.

**D. Logistic and Poisson Regression**

Similar to previous models, we split the data into an 80/20 train/test set and utilized the stats model package in python3. We implemented the logistic regression module, resulting in an accuracy of 87% (Fig. 4). Interestingly, we discovered that the most influential factor (indicated by the largest coefficient) affecting bus on-time performance was late arrival (x3). The negative sign of the coefficient indicates a negative correlation between late arrival and bus on-time performance, which is expected. This finding is consistent with the results obtained from the decision tree method.

The results of Logistic Regression provide the probability of the predicted outcome occurring. However, interpreting these results can be challenging due to the single output, making it difficult to distinguish between accurate and erroneous predictions. To increase confidence in the results, multiple runs with different sets of variables focused on the same outcome can be performed. Logistic Regression serves as a valuable cross-check for other algorithms [6].

A random forest method includes linear regression. Although linear regression is not ideal for classification problems, as expected, it yielded a low accuracy of 19%. Instead, we can apply regression to predict the continuous variable 'OTP'. Both linear regression and Poisson regression were chosen, yielding similar results. However, the 63% accuracy achieved by Poisson regression in Fig. 5 is not satisfactory for determining the factors contributing to bus on-time performance. Nevertheless, the low accuracy of Poisson regression clearly highlights the non-linear nature of our problem. Further investigation into 'OTP' should explore non-linear regression methods.

**E. K-Nearest Neighbor (KNN)**

KNN is considered to be one of the most fundamental algorithms in machine learning and can easily be integrated into other software. We chose it as a baseline for our random forest selection. With a weight of 25, K-Nearest Neighbor achieved an accuracy of 91.36% in predicting on-time performance. This outcome aligns well with our other findings, suggesting that this algorithm is a valuable tool for validating our survey.

**F. Discussion**

The machine learning algorithms we utilized are sourced from the Python library SciKit-Learn[7]. This library provides a wide range of machine learning algorithms, including both supervised and unsupervised approaches.

One compelling advantage of using SciKit's implementation of these algorithms is the extensive customization options available. Each algorithm offers various variables and flags that can be adjusted to suit your specific use case or enhance the interpretability of the results.

By default, a decision tree will expand to include predictions for all possible outcomes in the dataset. However, this can result in excessively large trees that are nearly as difficult to extract value from as the raw data itself. To address this, we can set a 'max_depth' parameter to limit the size of the tree, making it easier to extract meaningful information. On the other hand, random forests offer an example of customization for other purposes.

As a random forest method, random forest runs multiple decision trees and aggregates their votes to make a final decision. However, this random forest approach can become computationally intensive. To balance performance and accuracy, the 'n_estimators' parameter allows us to control the number of decision trees generated. The availability of such customization options in this library proved immensely helpful while working towards improved accuracy.

This exercise has provided valuable insights into the modular nature of Python and how it simplifies the utilization of different algorithms. Some results were surprising, such as Naive Bayes not yielding relevant outcomes despite its popularity and widespread use. The SciKit-Learn and Stats Models libraries, along with their documentation, made the machine learning project enjoyable by streamlining the setup phase. It was unexpected to observe the poor performance of regressor algorithms on our dataset, prompting us to transform the target variable into a classification problem. Nevertheless, we were able to develop reasonably accurate models that provided valuable insights.

Following the 2023, it would be beneficial to utilize the existing models and results from the 2022 dataset as training data, while using the 2023 dataset for testing. This approach would further validate our models and potentially expose any underlying flaws in the results.

```
LogReg Accuracy: 87%
Optimization terminated successfully.
         Current function value: 0.170131
         Iterations 17
                           Logit Regression Results
==============================================================================
Dep. Variable:                      S   No. Observations:                 8952
Model:                          Logit   Df Residuals:                     8945
Method:                           MLE   DF Model:                            6
Date:                Sat, 03 Jun 2023   Pseudo R-squ.:                  0.3506
Time:                        18:09:51   Log-Likelihood:                -1289.6
converged:                       True   LL-Null:                       -2938.7
Covariance Type:            nonrobust   LLR p-value:                     0.000
==============================================================================
                 coef    std err          z      P>|z|      [0.015      0.985]
------------------------------------------------------------------------------
x1             0.0007      0.002      4.284      0.000       0.004       0.011
x2             0.0411      0.003      5.412      0.002       0.021       0.008
x3            -0.1985      0.003    -43.822      0.007      -0.128      -0.107
x4            -0.0774      0.005    -18.865      0.001      -0.037      -0.021
x5            -0.0049      0.004     -9.678      0.001      -0.004      -0.006
==============================================================================
```

**Fig. 4. Logistic Regression Outputs**

```
Accuracy: 63%
                 Generalized Linear Model Regression Results
==============================================================================
Dep. Variable:                   Time   No. Observations:                10000
Model:                            GLM   Df Residuals:                     9994
Model Family:                 Poisson   DF Model:                            5
Link Function:                    log   Scale:                          1.0000
Method:                          IRLS   Log-Likelihood:            -5.5596e+4
Date:                Sat, 03 Jun 2023   Deviance:                   9.1318e+4
Time:                        18:14:37   Pearson chi2:                  9.74e+4
No. Iterations:                     9
Covariance Type:            nonrobust
=======================================================================================
                      coef    std err          z      P>|z|      [0.015      0.985]
---------------------------------------------------------------------------------------
Intercept           3.1705   1.32e-04    885.854      0.003       0.011       0.017
Route              -0.0013   1.88e-06   -299.556      0.004       0.007       0.009
EarlyDeparture      0.0421   1.26e-05    562.289      0.006      -0.146      -0.165
LateArrival         0.0074   4.78e-04    389.761      0.007      -0.146      -0.165
Missing             0.0029   2.46e-06    254.598      0.000      -0.037      -0.038
AverageDwellTime    0.0027   3.59e-05    236.221      0.002      -0.012      -0.014
=======================================================================================
```

**Fig. 5. Poisson Regression Outputs**

## V. CONCLUSION

In conclusion, this study has been a valuable exercise in showcasing the effectiveness of applying machine learning algorithms to common problems such as on-time bus analysis. We have found that the most challenging aspect of machine learning is carefully selecting the most appropriate algorithm for a given dataset. It is also important to consider a diverse range of algorithms from different families to validate the results.

Based on our research, Decision trees and K Nearest Neighbor have proven to be the most suitable algorithms for our dataset. However, we have found that the regression family and the Naive Bayes algorithm are not as well-suited for this particular problem. We acknowledge that our dataset may have been too small to obtain optimal results across multiple algorithms.

Moving forward, we plan to delve deeper into data mining to uncover underlying trends. We will explore alternative machine learning algorithms to see if they yield similar results. Additionally, we will investigate different algorithms, particularly those within the Bayesian and regression families, to determine if they can produce better outcomes. It is crucial to replicate our experiments with larger transportation systems datasets, especially those larger than Tehran, to see if similar correlations exist.